\documentclass{article}

\usepackage{PRIMEarxiv}

\usepackage[utf8]{inputenc} 
\usepackage[T1]{fontenc}    
\usepackage{hyperref}       
\usepackage{url}            
\usepackage{booktabs}       
\usepackage{amsfonts}       
\usepackage{nicefrac}       
\usepackage{microtype}      
\usepackage{lipsum}
\usepackage{fancyhdr}       
\usepackage{graphicx}       
\graphicspath{{media/}}     

\title{Synthetic Multimodal Dataset for Empowering Safety and Well-being in Home Environments
\thanks{\textit{The first and second authors contributed equally to this research.}} 
}

\author{
  Takanori Ugai \\
  Fujitsu Limited \\
  Kanagawa, Japan\\
  National Institute of Advanced Industrial\\ Science and Technology (AIST) \\
  Tokyo, Japan \\
  \texttt{ugai@fujitsu.com} \\
   \And
  Shusaku Egami\\
  National Institute of Advanced Industrial\\ Science and Technology (AIST) \\
  Tokyo, Japan \\
    \texttt{s-egami@aist.go.jp} \\
   \And
  Swe Nwe Nwe Htun \\
  National Institute of Advanced Industrial\\ Science and Technology (AIST) \\
  Tokyo, Japan \\
    \texttt{swenwe.nwehtun@aist.go.jp} \\
   \And
  Kouji Kozaki \\
  Osaka Electro-Communication University\\
  Osaka, Japan \\
  \texttt{kozaki@osakac.ac.jp} \\
   \And
  Takahiro Kawamura \\
  National Agriculture and Food Research Organization \\
  Ibaraki, Japan\\
  \texttt{takahiro.kawamura@naro.go.jp} \\
   \And
  Ken Fukuda \\
  National Institute of Advanced Industrial\\ Science and Technology (AIST) \\
  Tokyo, Japan \\
    \texttt{ken.fukuda@aist.go.jp} \\
}

\begin{document}
\maketitle

\begin{abstract}
This paper presents a synthetic multimodal dataset of daily activities that fuses video data from a 3D virtual space simulator with knowledge graphs depicting the spatiotemporal context of the activities.
The dataset is developed for the Knowledge Graph Reasoning Challenge for Social Issues (KGRC4SI), which focuses on identifying and addressing hazardous situations in the home environment.
The dataset is available to the public as a valuable resource for researchers and practitioners developing innovative solutions recognizing human behaviors to enhance safety and well-being in home environments.

\end{abstract}


\section{Introduction}
This paper introduces a multimodal dataset developed for the Knowledge Graph Reasoning Challenge for Social Issues (KGRC4SI)\footnote{\url{https://challenge.knowledge-graph.jp/2022/index_en.html}}. The challenge addresses the critical task of identifying hazardous situations in home environments using video data and/or knowledge graph reasoning techniques. 

The dataset combines simulated video data of daily activities with knowledge graphs (KGs) representing the spatiotemporal properties of the activities.
It includes 203 scenarios with 1,218 videos from different perspectives and room layouts, RDF-format~\cite{w3c2014rdf11} KGs of 2,902,676 triples, comprehensive schema details, and a SPARQL endpoint.

Four support tools were developed and are also made open to the public. Among them, two are discussed in this paper: VirtualHome-AIST, which triples the number of executable actions of the previous work, and VirtualHome2KG, a system that outputs simulation results as KGs. VirtualHome-AIST also generates 2D bounding box information for objects. VirtualHome2KG tracks object state changes and spatial situations, such as 3D coordinates.

The paper also discusses related works, highlighting scene graph generation and event-centered KGs. 

\section{Related Works}

In recent years, much attention has been paid to the study of scene graph generation (SGG), which represent video contents (scenes) in a graph structure~\cite{9661322}.
SGG data and our synthetic KG capture the video context, frame-by-frame, as a graph structure for the downstream tasks.
Qiu et al.~\cite{Qiu_2023_WACV}, discussing the label inconsistencies in existing datasets and ignorance of temporal relationships in SGG models, propose a synthetic dataset using VirtualHome~\cite{virtualhome} for SGGs. 
The novelty of our dataset is we incorporate Semantic Web technologies into the graph data management to strengthen the downstream reasoning tasks. It shall also be noted that our KGs are automatically generated from the information in the simulator and do not require manual annotations.

While KGs typically represent factual knowledge, KGs representing events such as historical events are known as Event Knowledge Graphs (EKGs)~\cite{guanWhatEventKnowledge2022}.
Several models have been proposed for EKGs, such as Simple Event Model (SEM)~\cite{van2011design} and Event Ontology\footnote{\url{http://motools.sourceforge.net/event/event.html}}.
Rospocher et al.~\cite{rospocher2016building} constructs EKG from news articles, and Gottschalk et al.~\cite{gottschalkEventKGMultilingualEventCentric2018} and Kawamura et al.~\cite{published_papers/3622142} construct EKGs from mystery novels. These EKGs are called Event-Centric KGs. This paper also adopts the Event-Centric KG model.

\section{Dataset Contents}

The dataset comprises Scenario Data, Video Data simulated from the scenario data using virtual space, and KGs representing the ``who,'' ``what action,'' ``object,'' ``when and where,'' and the resulting ``state'' and ``position'' of the objects. Knowledge Graph Embedding Data is provided to facilitate reasoning based on machine learning techniques. This dataset is available to the public as open data~\cite{ugai_takanori_2023_1204120}.

\subsection{Scenario Data}

\begin{table}
\caption{Number of Actions in Script Data}\label{table:Action}
\centering
    \begin{tabular}{rl|rl}
    Action & \# of Usage & Action & \# of Usage \\ \hline
SHAKE & 3 & SIT & 55 \\ \hline
CLOSE & 5 & WIPE & 97 \\ \hline
OPEN & 5 & PUTBACK & 105 \\ \hline
LOOKAT & 7 & SWITCHOFF & 106 \\ \hline
SCRUB & 22 & SWITCHON & 106 \\ \hline
EAT & 26 & TURNTO & 109 \\ \hline
PUTOBJBACK & 26 & GRAB & 298 \\ \hline
DRINK & 52 & WALK & 727
    \end{tabular}
\end{table}

\begin{table}
\caption{Number of Objects in Script Data}\label{table:Object}
\centering
    \begin{tabular}{rl|rl}
Object & \# of Usage & Object & \# of Usage \\ \hline
kitchen & 5 & pillow & 30 \\ \hline
television & 6 & desk & 32 \\ \hline
livingroom & 7 & alcohol & 39 \\ \hline
facecream & 12 & bread & 39 \\ \hline
toothbrush & 12 & cupcake & 39 \\ \hline
toothpaste & 12 & juice & 39 \\ \hline
groceries & 15 & milk & 39 \\ \hline
kitchencounter & 15 & wine & 39 \\ \hline
kitchentable & 15 & sponge & 50 \\ \hline
stove & 15 & sink & 51 \\ \hline
waterglass & 15 & bed & 54 \\ \hline
wallpictureframe & 21 & sofa & 66 \\ \hline
keyboard & 24 & washingsponge & 67 \\ \hline
mouse & 24 & towel & 114 \\ \hline
cat & 28 & remotecontrol & 290 \\ \hline
fridge & 28 & tv & 507
    \end{tabular}
\end{table}

The scenario comprises text files known as Script Data. VirtualHome2KG generates videos and knowledge graphs from this data as input. The Script Data include action titles and brief descriptions, all in text format. 
The 203 scripts use 16 different actions multiple times. The longest action sequence in the script data is 31, with a mean length of 8.6, a median length of 7, and a standard deviation of 5.6. Table \ref{table:Action} shows the number of each action in the data set. The total number of actions appearing in all scripts is 1749. The mean number of action occurrences is 109, the median is 52, and the standard deviation is 180.
Here, if an action, e.g., GRAB is included twice in one scenario, we consider the number usage of GRAB is 2.
GRAB and WALK are the most commonly used actions, especially WALK is included in every scenario. 59\% of the actions are accounted for by the two actions, Grab and Walk.
Table \ref{table:Object} shows the number of each object in the data set. The total number of objects appearing in all scripts is 1749. The mean number of object occurrences is 55, the median is 28, and the standard deviation is 97.
Here, if an object, e.g., tv is included thrice in one scenario, we consider the number usage of tv is 3.
Remotecontrol and tv are the most commonly used objects. 46\% of the actions are accounted for by the two objects, remotecontrol, and tv.
The scripts are grouped into six categories, 42 in EatingDrining, 7 in HouseArrangement, 70 in HouseCleaning, 70 in Leisure, 7 in SocialInteraction and 7 in Other. These categories follow the data set which VirtualHome provides.

\subsection{Video Data}
\begin{table*}
\caption{Number of Video in Categories}\label{table:Video}
\centering
    \begin{tabular}{c|cccccc|c}
 & {\scriptsize EatingDrinking} & \scriptsize HouseArrangement & \scriptsize HouseCleaning & \scriptsize Leisure & \scriptsize SocialInteraction & \scriptsize Other & \scriptsize All Categories \\ \hline
S1 & 36 & 6 & 66 & 60 & 6 & 6 & 180 \\ \hline
S2 & 36 & 6 & 60 & 60 & 6 & 6 & 174 \\ \hline
S3 & 36 & 6 & 54 & 60 & 6 & 6 & 168 \\ \hline
S4 & 36 & 6 & 54 & 60 & 6 & 6 & 168 \\ \hline
S5 & 36 & 6 & 60 & 60 & 6 & 6 & 174 \\ \hline
S6 & 36 & 6 & 66 & 60 & 6 & 6 & 180 \\ \hline
S7 & 36 & 6 & 60 & 60 & 6 & 6 & 174 \\ \hline \hline
All Scenes & 252 & 42 & 420 & 420 & 42 & 42 & 1218
     \end{tabular}
\end{table*}
The dataset contains videos in mp4 files derived from the 203 distinct action scenarios. Each scenario has various camera perspectives, including a rear view of the avatar (identified by file names ending in 0), an indoor camera with dynamic views (ending in 1), and fixed camera views positioned in each corner of the room (ending in 2-5). Furthermore, for each action scenario, we generated data representing a minimum of 1 and a maximum of 7 patterns, each depicting different room layouts(scenes). The dataset comprises a total of 1,218 videos. Table \ref{table:Video} shows the number of videos by scene and category.
The shortest video has 16 seconds, and the longest has 27 seconds.
Notably, specific videos intentionally simulate the movements of older adults with avatars moving slowly.

\subsection{Knowledge Graph}
\begin{figure}[t]
\centering
\includegraphics[width=1.0\columnwidth]{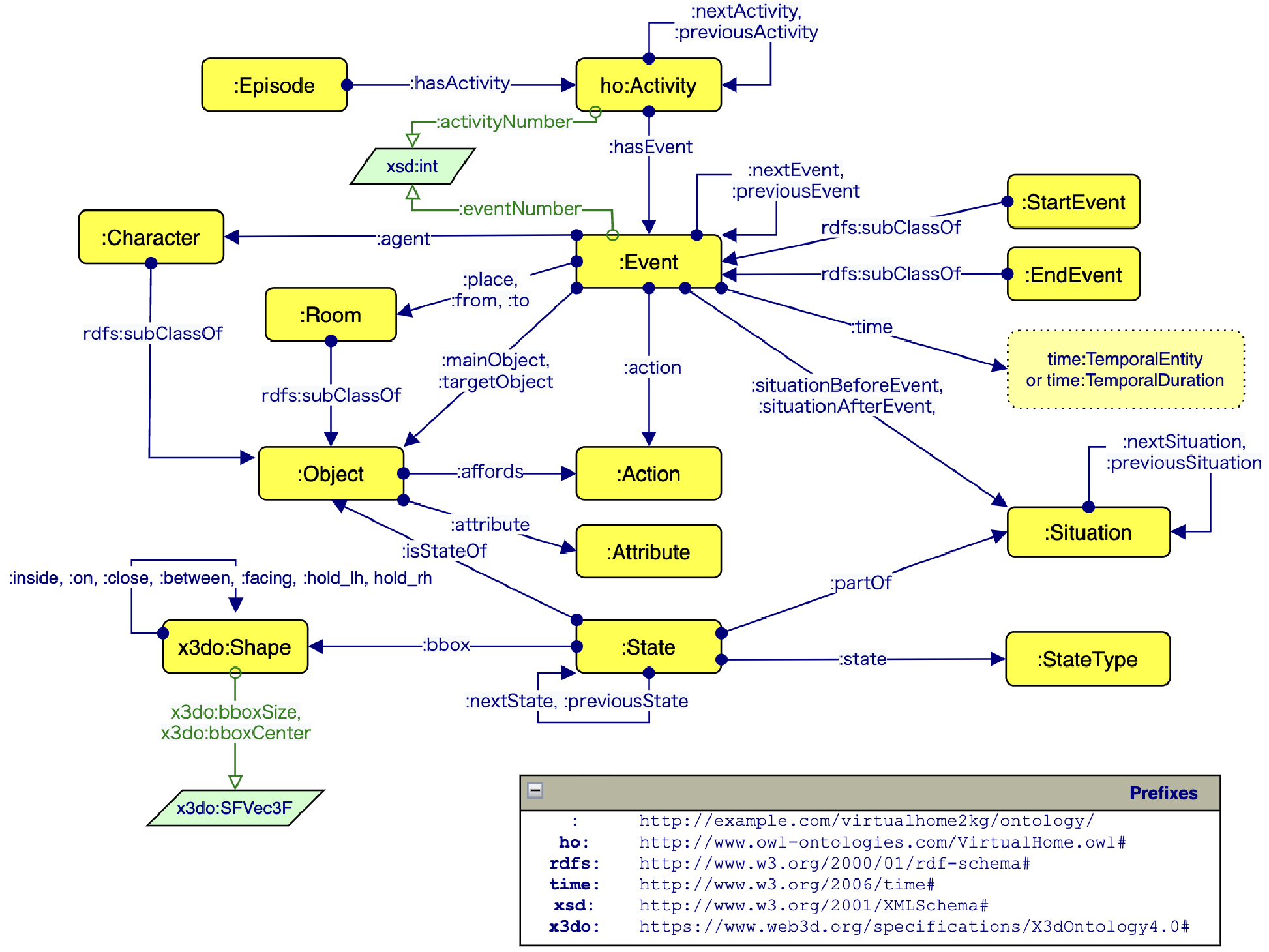}
\vspace{-2mm}
\caption{Class Diagram based on Event-Centric KG model~\cite{egami2023}}
\label{fig:schema}
\vspace{-5mm}
\end{figure}
The dataset includes KGs in RDF format. There are 203 knowledge graphs, each corresponding to the videos in the dataset, with 2,902,676 triples, 62 properties, and 772,358 entities excluding literals. One KG has an average of 17,195 triples. 
Additionally, the dataset offers SPARQL endpoint\footnote{\url{https://kgrc4si.ml:7200/sparql}} and query examples, enabling users to interact with the KGs.
Figure~\ref{fig:schema} shows how the Event-centric KG model represents the spatiotemporal information of the avatar and the objects.
An activity comprises multiple events where avatars, actions, objects, places, time, and situations are associated with each event.

\subsection{Embeddings}
The dataset incorporates embedding vectors generated using three models: TransE~\cite{NIPS2013_1cecc7a7}, ComplEx~\cite{10.5555/3045390.3045609}, and RotatE~\cite{sun2018rotate}. We created these embedding vectors using the DGL-KE~\cite{DGL-KE} with default parameters. Furthermore, we employed jRDF2Vec~\cite{DBLP:journals/corr/abs-2009-07659} to generate additional embedding vectors. The embedding data resulting from these approaches serve as a valuable resource for conducting machine learning-based reasoning and analysis.

\section{Supporting Tools}
\subsection{VirtualHome-AIST}

VirtualHome~\cite{virtualhome} is a Unity-based\footnote{\url{https://unity.com/}}  simulator with Python APIs for the simulator operation. 
It offers various 3D environments, representing different types of dwellings with diverse furnishings. 
The previous study implements 21 fundamental actions with motion data, as shown in Table \ref{table:action}\footnote{The table is based on VirtualHome ver.2.2}.
This study has implemented 37 new actions as VirtualHome-AIST to replicate a more comprehensive range of everyday scenarios, and 16 of them were used to synthesize the Scenario Data.
Since the cost of motion data implementation is high, the added actions were strategically selected from previous work that provides action classes as Primitive Action Ontology~\cite{nishimura2021ontologies}.

\begin{table*}[t]
\caption{Actions already implemented in VirtualHome and actions we have added}\label{table:action}
\begin{tabular}{|c|p{9cm}|}
\hline
Motion implemented actions in VirtualHome & Find, Walk, Run, WalkTowards, WalkForward, TurnLeft, TrunRight, Sit, StandUp, Grab, Open, Close, PutBack, Put, PutIn, SwitchOn, SwitchOff, Drink, TurnTo, LookAt, Touch                                                                                                  \\ \hline
Actions with added motion         & Wipe, Read, Pour, Type, Squeeze, Cut, Eat, Brush, Fold, Jump, JumpUp, JumpDown, Kneel, Lift, Rinse, Squat, Streach, Sweep, Stir, Throw, Unfold, Vacuum, Wrap, Write, Fall, FallSit, FallTable, FallBack, Climb, GoDown, Stand, Straddle, LegOpp, Scrub, Shake, Smell, Soak \\ \hline
\end{tabular}
\end{table*}

\subsection{VirtualHome2KG}

\begin{figure}[t]
\centering
\includegraphics[width=8.5cm]{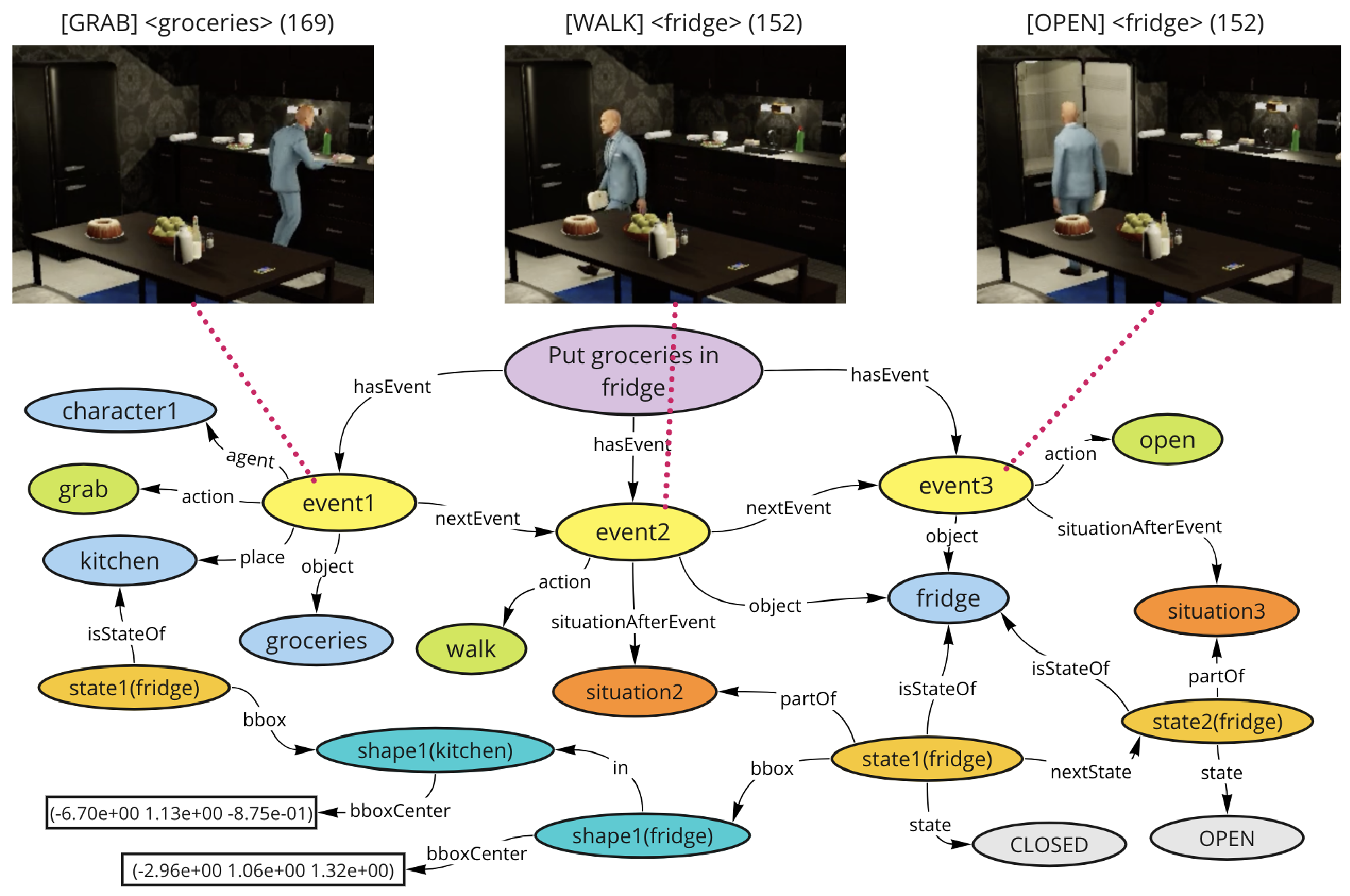}
\vspace{-2mm}
\caption{Event-centric KG and corresponding video frames}
\label{fig:kg_and_image}
\vspace{-5mm}
\end{figure}

VirtualHome2KG~\cite{ictai2021egami,egami2023} extends VirtualHome to store the detailed context information as KGs. 
%
In VirtualHome2KG, the activity is a ``live'' activity. For example, in Figure~\ref{fig:kg_and_image}, ``[GRAB] \verb|<|groceries\verb|>| (169)'', ``[GRAB] \verb|<|fridge\verb|>| (152)'', ``[WATCH] \verb|<|fridge\verb |>| (152)'', represent a part of a sequence of actions, object names, and object IDs contained in the activity named ``Put groceries in fridge''.
And when this is executed, VirtualHome and VirtualHome2KG update and record the environment information of the virtual space.
After ``[WALK] \verb|<|fridge\verb|>| (152)'' is executed, the coordinates of the avatar and the groceries are updated, and after ``[OPEN] \verb|<|fridge\verb|>| (152)'' is executed, the fridge state is updated to ``OPEN''. Execution of each line of the script data results in an Event node with all the auxiliary information stored around the Event node (Figure~\ref{fig:kg_and_image}). This recording mechanism captures the indoor environment at every step of the activity execution.

\subsection{Visualzation Tool and Scenario Editor}
We developed a visualization tool\footnote{\url{https://github.com/aistairc/virtualhome2kg_visualization}} to synchronize and display videos and KGs.
The tool plays the chosen video and enumerates the events within the video while concurrently visualizing and zooming in to the corresponding KG to support the comprehension of the video's event structure.
We also developed a scenario editor\footnote{\url{https://github.com/aistairc/virtualhome2kg_generation}} to assist in writing script data.
Creating an executable script is tedious as one must choose combinations of feasible movements and objects. Even if the script runs successfully, one has to assess if the resulting motion of the avatar is not unnatural.
This tool streamlines the data creation process by showcasing previous successful scripts.

\section{KGRC4SI}
The Knowledge Graph Reasoning Challenge for Social Issues (KGRC4SI) is a dedicated challenge for systems to identify and explain hazardous situations in home environments. The first challenge focused on older adults.
We envision detecting accident risks, such as falls, caused by human behavior or human-environment interaction.
By providing a multimodal dataset of visual information and structured symbolic information, we expect the participation of experts from diverse fields and the dataset to be a platform for applying technologies that aim to understand humans' daily behaviors.

\subsection{Challenge Tasks}

We defined one main task and two subtasks for tools and metrics. 
Here, we focus on the main task.
The main task comprises three smaller tasks:
\begin{enumerate}
\item {\bf Detect dangerous situations by analyzing the provided dataset.}
Identify environments, actions, and combinations that pose a high risk of accidents for older adults. The output should include frame IDs of the videos and partial graphs of the KGs.
\item {\bf Explain the reasons behind the identified risks.}
Describe the actions of the avatars, the involved objects, and their relationships.
\item {\bf Provide alternatives to decrease the detected risks.} 
Suggest an alternative activity that does not sacrifice the goal of the original risky actions.
\end{enumerate}

\subsection{Evaluation Criteria}

{\bf Main task}: 
The program committee (PC) verified the reproducibility of the submitted program code. 
The PC then rated the following items on a 5-point scale: 
(1) Adequacy assessment of defined risks, (2) Detail degree of defined risks, (3) Accuracy evaluation of detected risks, (4) Explanation's quality of detected risks, (5) Goal alignment of alternative solutions, (6) technical innovations and knowledge resource contribution.

\subsection{Accepted Entries}
We hosted KGRC4SI 2022 in Japan as a pilot project with a six-month application period with 6 accepted submissions.

Five of the six entries had publicly available sources and data, and the PC member could also apply the private data prepared by the contenders.
Diverse technologies were adopted in the entries, including knowledge processing, knowledge graph embedding, and generative AI.
An example of the identified risk was ``Trying to reach for something in a place higher than oneself.''
Three entries applied rule-based approaches to delineate risks and applied knowledge processing techniques to identify potential risks, while two entries applied link prediction with knowledge graph embedding. Three entries utilized Large Language Models (LLMs) to facilitate description generation, thereby enabling the detailed explanation of risks and the generation of alternative solutions. One entry used generative AI to identify risks within videos.
Two entries used generative AI by LLMs to address the subtask of script creation.

\section{Summary and Future Works}
The paper introduces a synthetic multimodal open dataset of daily human activities, merging visual information (video) and structured symbolic information (KG). Four systems have been made available to the public to generate the dataset. One system triples the number of avatars' activities compared to previous work, and the other enables automatic KG generation as the gold standard annotation of the video content.
Compared to the previous SGG datasets, the novelty of the dataset is that we incorporate Semantic Web technology to manage the intrinsically complex relationships among objects and make them reusable for both machine learning and symbolic reasoning approaches, which is impossible for the current SGG datasets.
Through the pilot challenge, we discovered that our dataset challenges LLMs. While LLMs do provide general know-how answers based on common sense, the dataset is unseen to the LLM model and requires a grounding of the physical world to some extent to provide case-based concrete solutions.

The current dataset only has RDF-based KGs, and we plan to provide the graph data in \textit{de facto} SGG format for easy handling in machine learning applications. We plan to add more scenarios, videos, and graph data continuously. By doing so, we expect the dataset platform becomes applicable to other domains, including children, factory workers, and traffic safety.


\section*{Acknowledgement}
This paper is based on results obtained from a project, JPNP20006, commissioned by the New Energy and Industrial Technology Development Organization (NEDO), and Japan Society Promotion of Science (JSPS) KAKENHI Grant Numbers JP19H04168, JP22K18008, and JP23H03688.

\bibliographystyle{unsrt}
\bibliography{main}

\end{document}